\documentclass[]{spie}  

\usepackage{amsmath,amsfonts,amssymb}
\usepackage{graphicx}
\usepackage[colorlinks=true, allcolors=blue]{hyperref}
\usepackage{amssymb}
\usepackage{rotating}
\usepackage{float}
\usepackage {physics}
\usepackage{amsmath}
\usepackage{yfonts}
\usepackage{csquotes}
\usepackage[english]{babel}
\usepackage{graphicx}
\usepackage{hyperref}
\usepackage[nottoc,notlot,notlof]{tocbibind}
\usepackage{mathtools}
\usepackage{wrapfig}
\usepackage{notes2bib}
\usepackage{authblk}
\usepackage{subcaption}
\usepackage[font = small]{caption}

\title{XNet: A convolutional neural network (CNN) implementation for medical X-Ray image segmentation suitable for small datasets}

\author[]{Joseph Bullock}
\author[]{Carolina Cuesta-L\'{a}zaro}
\author[]{Arnau Quera-Bofarull}
\affil[]{Department of Physics, Durham University, UK}

\authorinfo{Further author information: (Send correspondence to Joseph Bullock)\\Joseph Bullock: E-mail: j.p.bullock[at]durham.ac.uk, \\  
Carolina Cuesta-L\'{a}zaro: E-mail: carolina.cuesta-lazaro[at]durham.ac.uk, \\
Arnau Quera-Bofarull: E-mail: arnau.quera-bofarull[at]durham.ac.uk}

\begin{document} 
\maketitle

\begin{abstract}
      X-Ray image enhancement, along with many other medical image processing applications, requires the segmentation of images into bone, soft tissue, and open beam regions. We apply a machine learning approach to this problem, presenting an end-to-end solution which results in robust and efficient inference. Since medical institutions frequently do not have the resources to process and label the large quantity of X-Ray images usually needed for neural network training, we design an end-to-end solution for small datasets, while achieving state-of-the-art results. Our implementation produces an overall accuracy of 92\%, F1 score of 0.92, and an AUC of 0.98, surpassing classical image processing techniques, such as clustering and entropy based methods, while improving upon the output of existing neural networks used for segmentation in non-medical contexts. The code used for this project is available online \cite{Code}.
\end{abstract}

\keywords{X-Ray image segmentation, medical image processing, machine learning.}

\section{Introduction}\label{Introduction}


X-Ray image segmentation is of great importance in many medical applications such as image enhancement, and other processing tasks, computer assisted surgery, and anomaly detection. These applications regularly require the segmentation of images into 3 categories: open beam, soft tissue and bone
. Current methods rely heavily on a complex system of classical image processing techniques, such as clustering approaches \cite{Clustering_Approach, Spectral_Clustering}, line fluctuation analysis \cite{Line_Fluctuations}, or entropy-based methods \cite{Entropy_1, Entropy_2}, which often require the tuning of hyperparameters for each body part class. However, utilising machine learning offers several advantages over these traditional methods since: (i) it naturally addresses noise, (ii) it generalises well to different body parts, and (iii) the segmented regions have continuous boundaries.

Although the use of machine learning in the healthcare sector has grown significantly in recent years, and with it open source datasets have become more readily available \cite{ChestX-Wang, MURA, DeepLesion, Preparing_a_Collection}, we know of no publicly available labelled dataset of X-Ray images that can be used in training a neural network for segmentation tasks as presented in this paper. This means institutions wishing to train such a network must provide and label their own images. X-Ray images are expensive to obtain, and manual labelling is time consuming, thus acquiring a large and varied database may not be possible; a common issue raised when discussing the feasibility of neural networks for X-Ray segmentation \cite{Review_1}.

We design a unique Convolutional Neural Network (CNN) architecture to perform segmentation by extracting fine grained features, while controlling the number of trainable parameters to prevent overfitting. The network is trained on 150 X-Ray images, with no scatter correction and comprising of 19 body parts in an imbalanced way. This dataset, we believe, is of a manageable size to be created by a medical institution. We present our network, and a full end-to-end description of it's implementation, including post-processing stages for the minimisation of false positives, and optimisation of the F1 score. Despite our dataset being small compared to those used in many machine learning applications, we achieve an overall accuracy significantly higher, and more generalisable, than work using classical image processing techniques \cite{Clustering_Approach, Spectral_Clustering, Line_Fluctuations, Entropy_1, Entropy_2, Atlas}. Additionally, we show that our architecture outperforms leading image segmentation networks developed for other applications \cite{SegNet_Basic}.

Our paper is structured as follows: after reviewing a selection of the existing literature in Section \ref{Related}, we discuss our dataset in Section \ref{Data} - how we collect and label the data, and the augmentation methods to prevent overfitting; in Section \ref{Architecture} we address the design and structure of our CNN, covering training and testing stages; the results of the network are discussed in Section \ref{Results}, where we also address the post-processing stage of false positive reduction;  a comparison of our results with other works from the classical and machine learning literature is presented in Section \ref{Conclusion}, after which we discuss future applications and developments in Section \ref{Future}.

\section{Related Work}\label{Related}


Note that we provide this section not as an extensive literature review, but to lay the groundwork for where the methodology and results of this work sit in the current research landscape. We include work from which we have drawn important insights and information, while attempting to provide a pedagogical introduction to the development of the field of X-Ray image segmentation, along with modern image segmentation methods in a broader context.

Methods for segmenting X-Ray images have been a constant topic in the literature for many years, due to its role in image processing and other analysis based operations. Much previous work focuses on using classical image processing techniques \cite{Review_1, Review_2}. Pixel clustering based on similarity in certain parameters is a commonly employed technique. Kubilay Pakin \textit{et al.} \cite{Clustering_Approach} use clustering as a component of their segmentation algorithm, achieving high accuracy scores, but requiring hyperparameter tuning for each body part class, while struggling to produce smooth boundaries. Similarly, good results have been achieved by Wu and Mahfouz \cite{Spectral_Clustering}, who employ spectral clustering methods and produce much smoother boundaries, however, this application was tuned purely to knee image analysis. Entropy-based approaches have been employed by Bandyopadhyay \textit{et al.} \cite{Entropy_1, Entropy_2}, which have enabled clean boundary identification, yet suffer from extraneous edges such as bone cracks or image distortions. Kazeminia \textit{et al.} \cite{Line_Fluctuations} build on this work, employing existing edge detection algorithms such as \textit{Sobel} \cite{Sobel} and \textit{Canny} \cite{Canny}, while analysing the intensity fluctuations in pixel rows to more accurately select the bone boundary. This work is less sensitive to noise, however, loses some boundary continuity. Similarly, atlas models have been developed for medical image segmentation, and an application to rib cage segmentation is given by Candemir \textit{et al.} \cite{Atlas}. From such techniques the authors are able to generate complete segmentations, however, the boundaries remain noisy and the area under the ROC curve (AUC) is highly dependent on the dataset analysed.


The growth of machine learning applications in the healthcare sector has been considerable in recent years. Indeed, due to the large amount of information encoded in X-Ray images, focused research into their analysis has been significant. Aiding this advancement of research has been an increase in availability of X-Ray image datasets, each tailored to different applications \cite{ChestX-Wang, MURA, DeepLesion, Preparing_a_Collection}. Using such datasets, many machine learning approaches for the detection of anomalies, such as pneumonia \cite{CheXNet}, pulmonary tuberculosis \cite{RSNA}, and thoracic diseased \cite{Thoractic_Diseases} have been developed, as well as \textit{diagnosing} a variety of diseases based on chest X-Rays \cite{Learning_to_Diagnose}. Islam \textit{et al.}\cite{Abnormality} and Qin \textit{et al.}\cite{AI_Review} provide comparisons of a range of neural networks applied to detecting anomalies in chest X-Rays. These detection technologies usually perform image segmentation to localise the position of the anomaly, or of a certain bone structure (most commonly the rib cage), and are carefully tuned to these applications. Additionally, due to their architectures, many of the networks presented in the above literature would not be suitable for performing a pixel-level segmentation over the entirety of the image.



Complete multi-class image segmentation is an active, high-growth area of research, most recently driven by autonomous vehicle development. Neural networks have been at the forefront of this research and take various forms including the encoder-decoder design similar to that presented in this paper \cite{SegNet, DDNN, Learning_Deconvolution, Conditional_Random_Fields, UNet}, and fully connected networks \cite{FCN}. Several applications of these networks also utilise the technique of image augmentation to aid network generalisation and reduce the risk of overfitting. Similarly, Badrinarayanan, Handa and Cipolla \cite{SegNet_Basic} present a simplified version of the widely applied SegNet architecture\cite{SegNet}, showing improved performance on small datasets. Indeed, it is against this simplified network, known as \textbf{SegNet-Basic} \cite{SegNet_Basic}, that we benchmark our network performance. It is noted that some networks have been specifically designed for the total segmentation of medical images, yet these applications have been largely constrained to the segmentation of cell structures \cite{UNet, DNN_NIPS, Competition}. There have been examples of networks, such as U-Net \cite{UNet}, being applied to other segmentation tasks in the field of X-Ray image segmentation \cite{Use_of_UNet}, but only in specific use cases.

\section{Data}\label{Data}

\subsection{Collection and Labelling}

We use data collected from two sources at IBEX Innovations Ltd. \cite{IBEX}: 69 CT scans images of feet, knees and phantom heads, and  81 standard X-Ray images of different body and phantom body parts, of which the thorax is the most underrepresented. Several images contain foreign metal objects, such screws and staples, which we aim to classify as bone. The particular distribution among body part classes can be seen in Table \ref{table}. The images have not been corrected for scatter effects; but they have been dark-corrected by removing the detector background signal. Additionally, we pre-process each image by performing mean subtraction and pixel value normalisation to within the range $[-1,1]$.  The images are labelled using the free software \textsc{Gimp} \cite{GIMP}, by assigning a colour to each of the three distinct regions: open beam, soft tissue and bone. As a case study we specifically aim to minimise the soft tissue false positives, and so primarily avoid labelling any bone or open beam region as soft tissue. Both the images and labels are resized to $200 \times 200$ pixels to match the network input shape.

\begin{center}
\begin{tabular}{ |p{0.25\linewidth}|p{0.1\linewidth}||p{.25\linewidth}|p{.1\linewidth}| }
 \hline
 \multicolumn{4}{|c|}{Bodypart list} \\
 \hline
 \hline
 Ankle & 10 & Leg & 1\\
 \hline
 Arm & 3 & Lumbar spine & 6 \\
 \hline
 Cervical & 1 & Neck of femur & 15 \\
 \hline
 Chest & 1 & Pelvis & 2 \\
 \hline
 Elbow & 1 & Shoulder & 2 \\
 \hline
 Femur & 3 & Thigh & 8 \\
 \hline
 Foot & 29 & Thorax & 1 \\
 \hline
 Hand & 4 & Tibia & 4 \\
 \hline
 Head & 11 & Wrist & 11 \\
 \hline
 Knee & 36 & \textbf{Total:} & \textbf{150} \\
 \hline
\end{tabular}
\vspace{0.3cm}
\captionof{table}{Number of images in each body part class on our dataset. Note that the different body part classes are highly unbalanced.}
\label{table}
\end{center}

\subsection{Augmentation}

Compared to previous successful applications of neural networks to image segmentation, our dataset is small. There are 150 images in total, with unbalanced body part classes, representative of that which may be created by a medical institution. Therefore, we artificially augment the training images with the two-fold purpose of creating a larger dataset, to avoid overfitting, and balancing the different body part classes through augmented oversampling. After experimenting with a variety of filters, we find that elastic transformations are a crucial component for generating realistic augmented images, combined with translations, rotations, shear and cropping. 
For our dataset, we also find augmenting to produce 500 images per body part class gives the highest validation accuracy. If there is only one example of a particular body part, this image is placed in the test set and not used for training. 

\section{Architecture}\label{Architecture}
\begin{figure*}[!htbp]
    \centering
\includegraphics[width = \linewidth]{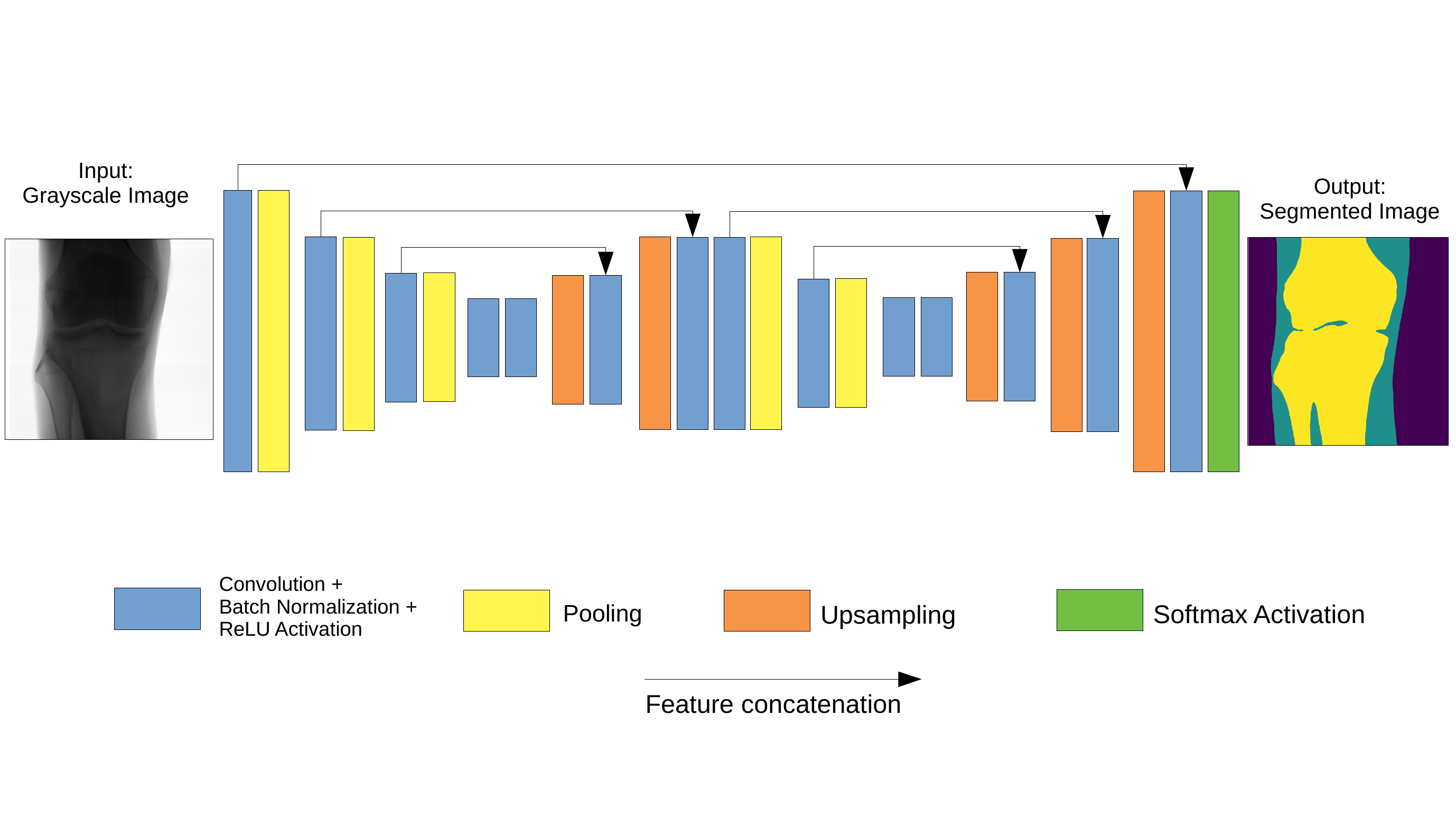}
\vspace{-1cm}
\caption{Visualisation of XNet architecture including example input image, left, and output segmented mask, right. Feature concatenation of same dimension layers helps to avoid losing fine-grained detail. Softmax activation function provides final pixel-wise classification.}
\label{architecture}
\end{figure*}
\subsection{XNet}


XNet is based on an \textit{encoder} - \textit{decoder} style architecture commonly used in image segmentation \cite{SegNet, DDNN, Learning_Deconvolution, Conditional_Random_Fields, UNet}.

\textbf{Encoder} The encoder consists of a series of convolutional layers, for feature extraction, and max pooling layers to downsample the input image. Breaking up the downsampling into a multiple stages allows for varying levels of extraction, with increasingly global features learnt through the convolutional layers at each
pooling stage.

\textbf{Decoder} After feature extraction, the decoder performs upsampling to generate a segmented mask of equal dimension to the input image. Similar to the encoder, using a multistage upsampling process with convolutional layers in between allows for varying degrees of fine grain feature reconstruction during upsampling, thus producing dense feature maps.

 

Due to the small size of our dataset, we avoid large \textit{serial} downsampling of the input image compared to many other networks, particularly those used in image classification. We avoid this since performing a greater number of downsamplings in series can be detrimental to accurate boundary level detail, particularly around smaller structures. However, downsampling allows for learnable feature extraction, and so is important to include in the network. 

We present an architecture which incorporates a comparable, or greater, number of downsampling stages for feature extraction as other segmentation networks, whilst avoiding overly-reducing image resolution. This is achieved by using two encoder-decoder modules in succession, whilst storing encoder feature maps and using them during the creation of the dense feature maps in the decoders (as can be seen in Figure \ref{architecture}).

Each convolutional layer is associated with a rectified-linear non-linearity (ReLU) activation function, and the decoder upsamples the image using nearest-neighbours upsampling. Storage and use of the encoder feature maps is performed through filter copying between layers of equivalent dimensions, meaning the model is less likely to `forget' what it has previously learnt, thus decreasing the likelihood of loosing fine-grained detail after performing downsampling (a technique used in several encoder-decoder networks \cite{SegNet, UNet}). At each convolutional layer we employ L2 norm regularisation, with penalty parameter $\lambda = 5\times10^{-4}$, to improve network generalisation and prevent overfitting \cite{Weight_Decay}. For more detailed information regarding network architecture and parameters, including filter sizes, see the supplementary material \cite{Code}.


\subsection{Training}

We train our model on augmented data, thus increasing the number of training examples, thereby reducing the chance of overfitting. Of the 150 images in our original dataset, 108 are set aside for augmentation and training. The remaining images are used in validation and testing stages in equal proportion.

The ground truth masks used for training are one-hot encoded. For each input pixel $X$ we optimise the categorical cross-entropy loss,
\begin{equation}
L(X,y) = -\sum_{i}\,I(y,i)\,\text{log}\,p(Y = i|X),
\end{equation}
where $y$ is the output label generated by the network, $p(Y = i|X)$ is the probability that the network assigns the label $i$ given the input data, and $I(y,i)$ is the indicator function defined by

\begin{equation}
I(y,i) = 
\begin{cases}
 0 & \text{if} \,\,\, y\neq i,\\
 1 & \text{if} \,\,\, y=i.
\end{cases} 
\end{equation}

We train using Adam optimisation \cite{Adam} with learning rate $10^{-4}$.

Since our aim is to design a network that could also be retrained by a non-specialist institution who may not have access to large memory GPUs, we choose small mini-batch and kernel sizes. We train on a batch size of 5 with each convolutional layer having a kernel size of $3\times 3$. To optimise training time we use \textit{Early Stopping} (see Section 7.8 of Goodfellow, Bengio and Courville \cite{Deep_Learning_Book}) by monitoring `validation loss' with a patience of 20. The validation set is not augmented and chosen by randomly selected at least one image from each body part class, so as to avoid tuning to a bias dataset.
Training using the above parameters took 7 hours on a GTX 1060 6GB GPU.

\subsection{Testing}

We test our network on images without augmentation chosen in a similar way to the validation set. Indeed, we manually classify certain images as `difficult' based on factors such as: bone structure complexity, noise, and the contrast ratio. To ensure our network can handle such difficulty, and generalise well, we ensure that the test set contained a disproportionate number of these difficult cases.
We carry out all testing before post-processing, and the accuracy score is calculated using the \textit{categorical accuracy} metric in Keras \cite{Keras}.

Although we correct for body part class imbalance at the pre-processing stage, our dataset is still highly imbalanced in the segmentation categories. The open beam region is both the most prevalent in the dataset and the easiest to classify (during architecture development and hyperparameter tuning this category always achieved the highest accuracy). Therefore, accuracy over all classes is not necessarily the most effective metric by which to measure the model's performance.

For our applications we look for a balance between false positive reduction and true positive enhancement. Therefore we use the F1 score as a measure of network performance. The comparison of accuracy and F1 scores can be found in Section \ref{Results}.






\section{Results}\label{Results}
\subsection{Network Performance}

\begin{figure}[h!]
  \begin{subfigure}[t]{.45\textwidth}
  \centering
  \hspace*{-1cm}%
    \includegraphics[width=1.2\textwidth]{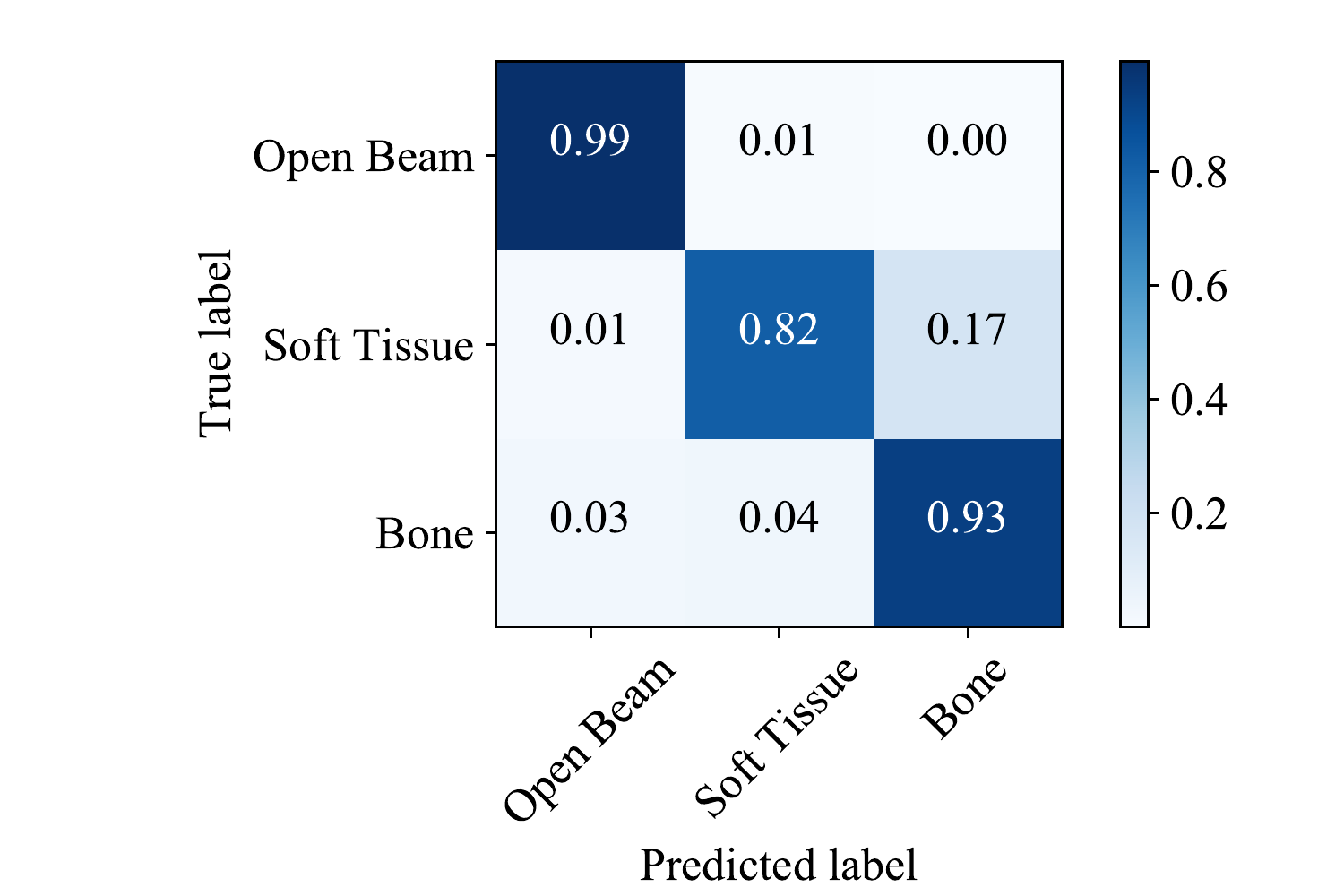}
    \caption{Confusion matrix}
    \label{confusion}
  \end{subfigure}\hfill
  \begin{subfigure}[t]{.45\textwidth}
  \centering
    \includegraphics[width=1\textwidth]{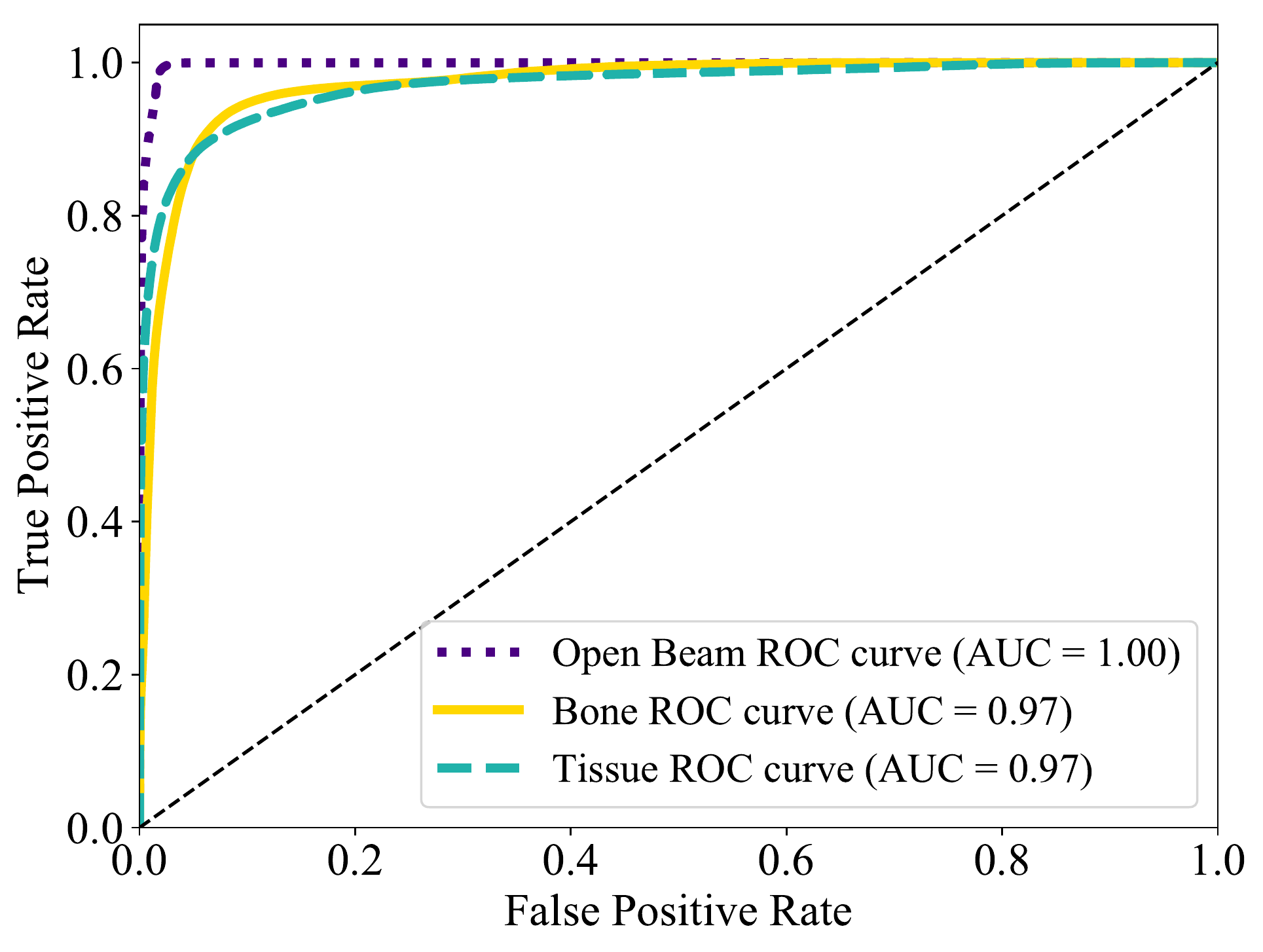}
    \caption{ROC curve}
    \label{roc}
  \end{subfigure}
  \vspace{0.1cm}
  \caption{(a) Normalised confusion matrix of XNet results. Each row represents the instances in a class, while each column shows the class predicted by the network for those instances. Here, most of the errors are made classifying soft tissue as bone. (b) ROC curve showing the true positive rate against the false positive rate for the open-beam, bone and soft-tissue categories. The area under the curve (AUC) for the different categories is shown. \vspace{1cm}}
\end{figure}

\begin{figure*}[h!]
\begin{center}
\includegraphics[width = 1.\linewidth]{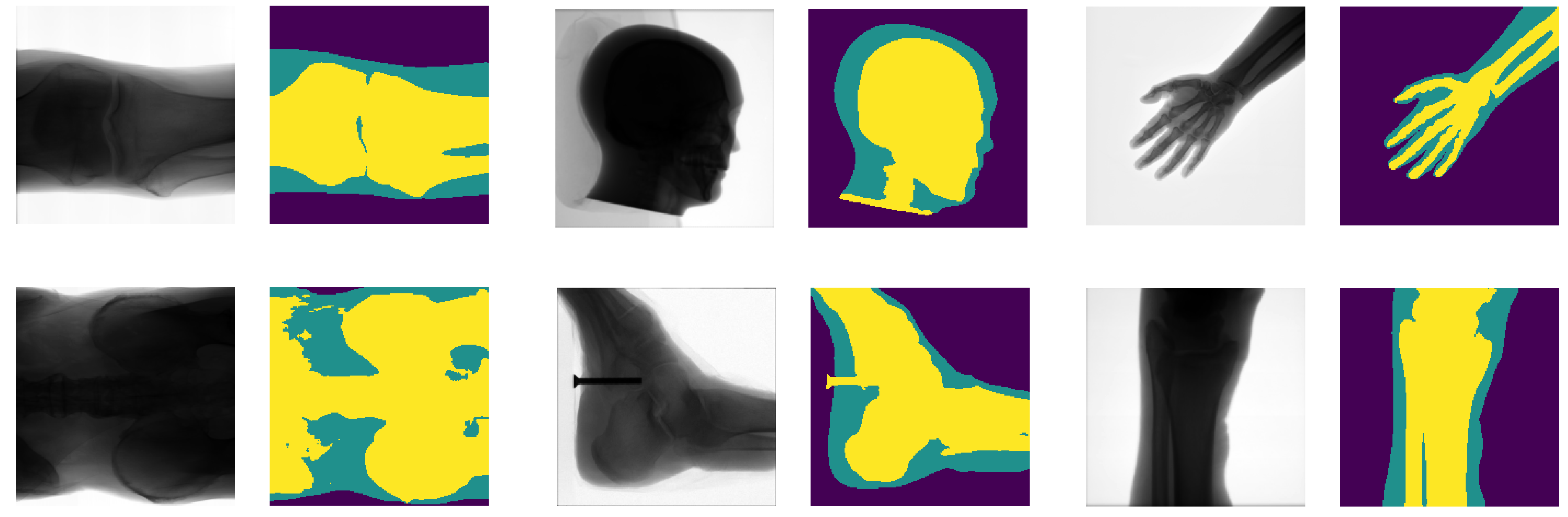}
\end{center}
\caption{Segmentation predictions from the test set. Top row images show a knee, phantom head and hand. Bottom row shows a pelvis, ankle with a metallic bolt, and the lower half of a leg. The open beam area is shown in purple. Bone is shown in yellow and soft tissue in green.}
\label{predictions}
\end{figure*}

The network achieves an open beam, soft tissue, and bone classification accuracy of 96\%, 94\%, and 88\% respectively, and an overall weighted averaged accuracy of 92\%. 
Additionally, we obtain an F1 score of 0.97, 0.87, and 0.90, in the open beam, soft tissue, and bone categories respectively. Taking into account class imbalance, the overall weight averaged F1 score is 0.92. The summary of results is presented in Table \ref{table2}. In Figure \ref{predictions}, we show some predictions of images in our test set, with the confusion matrix computed over the test set presented in Figure \ref{confusion}. The network can be seen to perform well for different body parts, including on the more challenging, due to its complexity and absence of training examples, chest region. 

\begin{center}
 \begin{tabular}{|c c c c c|} 
 \hline
 Category & F1-Score & AUC & Accuracy & Confidence\\ [0.5ex] 
 \hline\hline
 Open Beam & $0.97$ & $1.00$ & $96\%$ & $99\%$ \\ 
 \hline
 Soft tissue & $0.87$ & $0.97$ & $94\%$ & $95\%$\\
 \hline
 Bone & $0.90$ &  $0.97$ & $88\%$ & $97\%$\\
 \hline
 \textbf{Weighted average} & $\textbf{0.92}$ & $\textbf{0.98}$ & $\textbf{92\%}$ & $\textbf{97\%}$  \\
 \hline
 
\end{tabular}
\vspace{0.3cm}
\captionof{table}{Evaluation metrics for the three categories, and their weighted averages.}
\label{table2}
\end{center}

\subsection{Calibration}
Modern CNNs are often ill-calibrated, making them overconfident about their predictions \cite{calibration}. For networks broadly applicable to a variety of medical image segmentation tasks, such calibration error can be detrimental to important post-processing steps, such as false positive reduction in a given category. 
The output of XNet is a 3-dimensional probability map, where each pixel is assigned a probability of belonging to one of the different categories. We define network confidence in a given category as
\begin{equation}
    \text{conf}(X) = \sum_{i \in X} p_i, 
\end{equation}
where $X$ is the set of all pixels assigned to that category, and $p_i$ is the probability that the $i$th pixel belongs to said category. 

\begin{figure}[H]
    \centering
    \includegraphics[width = 0.8\linewidth]{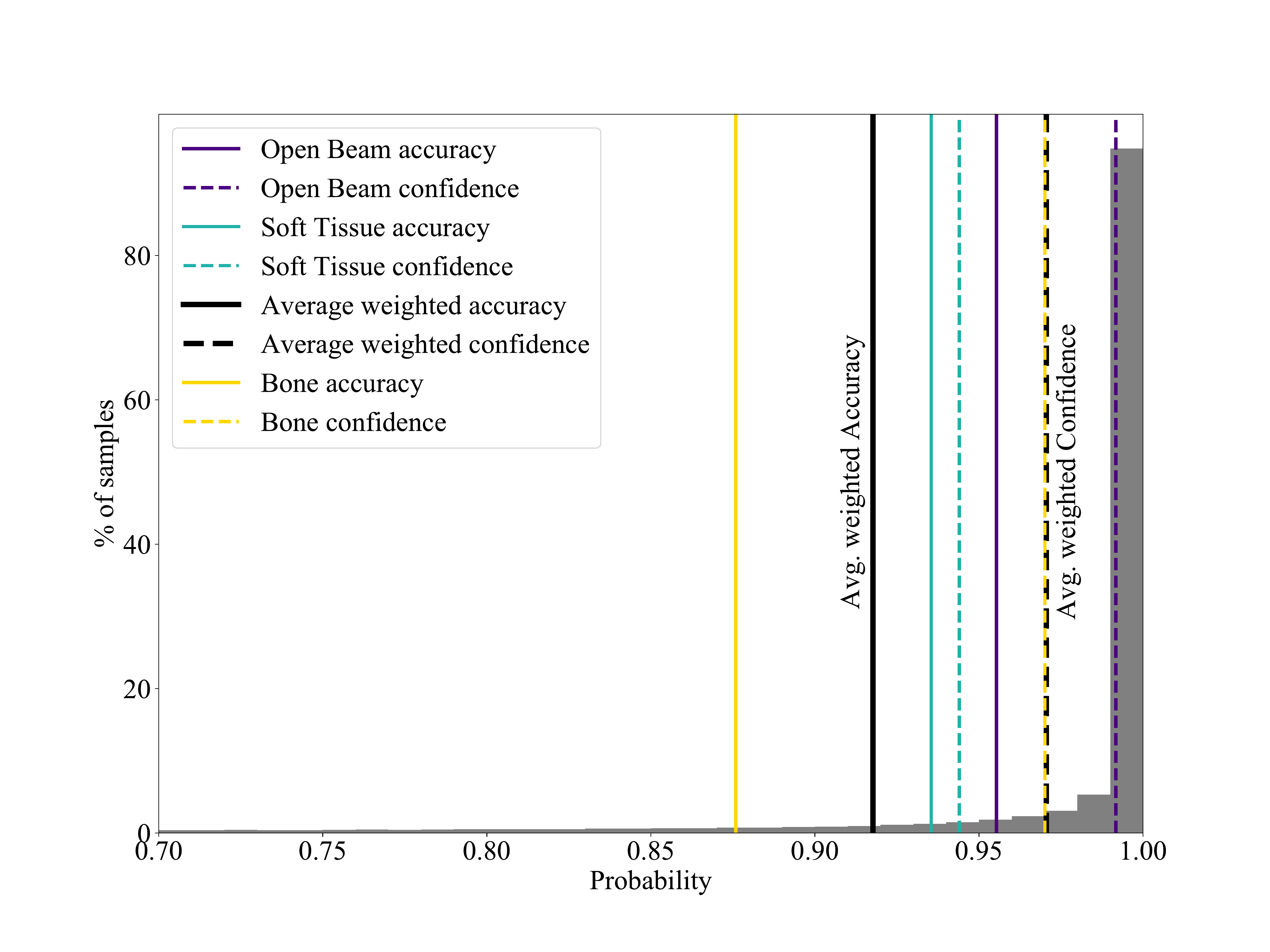}
    \caption{Confidence histogram across all categories, showing that the network is well calibrated. For the individual categories, the best calibrated is soft tissue with a difference of $1 \%$. For open beam the difference is $3 \%$. The least calibrated category is bone, with a difference of $9 \%$. Note: as mentioned, in our case study, when we carry out hyperparamter optimisation we are focusing on minimising soft tissue false positives, so these results are as expected.}
    \label{confidence}
\end{figure}

A well calibrated network is defined as a network whose confidence averaged over all output categories, is close to the network's averaged accuracy \cite{calibration}. We find that our network is indeed well calibrated, as defined by this metric, as can be seen in Figure \ref{confidence}. XNet is therefore well suited for the employment of a variety of post-processing techniques to tailor the output depending on the use case.


\subsection{False Positive Reduction}

The simplest way of obtaining a segmentation from the network output is to assign each pixel to the most likely category. Nonetheless, with our objective of reducing soft tissue false positives in the soft tissue category, we classify a pixel as soft tissue only if the network predicted probability is higher than a given threshold. Since our network is well calibrated, this probability threshold reflects the confidence that we demand for a pixel to be in this category. We find that increasing this probability threshold reduces the number of false positives at the expense of the number of true positives. For example, with a probability threshold of 0.90, the false positive rate in the soft tissue class is reduced to 3\%, while reducing the true positive rate to 71\%. Choosing the ideal probability threshold depends on the particular use case. We choose more severe thresholds for the body parts that the network struggles more to segment. 
We obtain an area under the curve of 1.00 for the open beam region and 0.97 for both the bone and soft tissue classes, thus demonstrating the high classification ability of our network.

\section{Conclusion}\label{Conclusion}



We develop a fully automatic method to segment medical X-Ray images given a small dataset. As a compromise between having a deep network to extract high-level features and fine-grained detail, and a network that avoids overfitting to small datasets, we present an architecture with two encoder-decoder modules. We train this network on a dataset consisting of 150 images, artificially augmented to generate 7000 training images. Evaluating the network performance we find an overall accuracy of 92\% and F1 score of 0.92, with an AUC of 0.98.

We benchmark our network against the popular SegNet design. Starting with the Segnet-Basic architecture \cite{SegNet_Basic}, and carrying out a hyperparameter search similar to the one we do for XNet, the best result obtained gives a 2\% improvement on bone classification accuracy with respect to XNet, but at the expense of only having a 75\% true positive rate and 25\% false positive rate in the soft tissue category (in comparison to our network which achieves an 82\% TP ratio, see Figure \ref{confusion}). Additionally, we significantly outperform this network when detecting the open beam region. SegNet-Basic achieves F1 scores of 0.96, 0.83, and 0.90 for open beam, soft tissue and bone regions respectively, with an overall weighted F1 score of 0.89, thus giving a lower F1 score in every category compared to XNet (see Table \ref{table2}). The full implementation of SegNet \cite{SegNet} has too many parameters to effectively fit to such a small dataset. Similarly, its size requires significantly more computational power, thus making it impractical for the purpose of being trainable by many medical institutions.

\begin{figure}[H]
\begin{center}
\includegraphics[width = .4\linewidth]{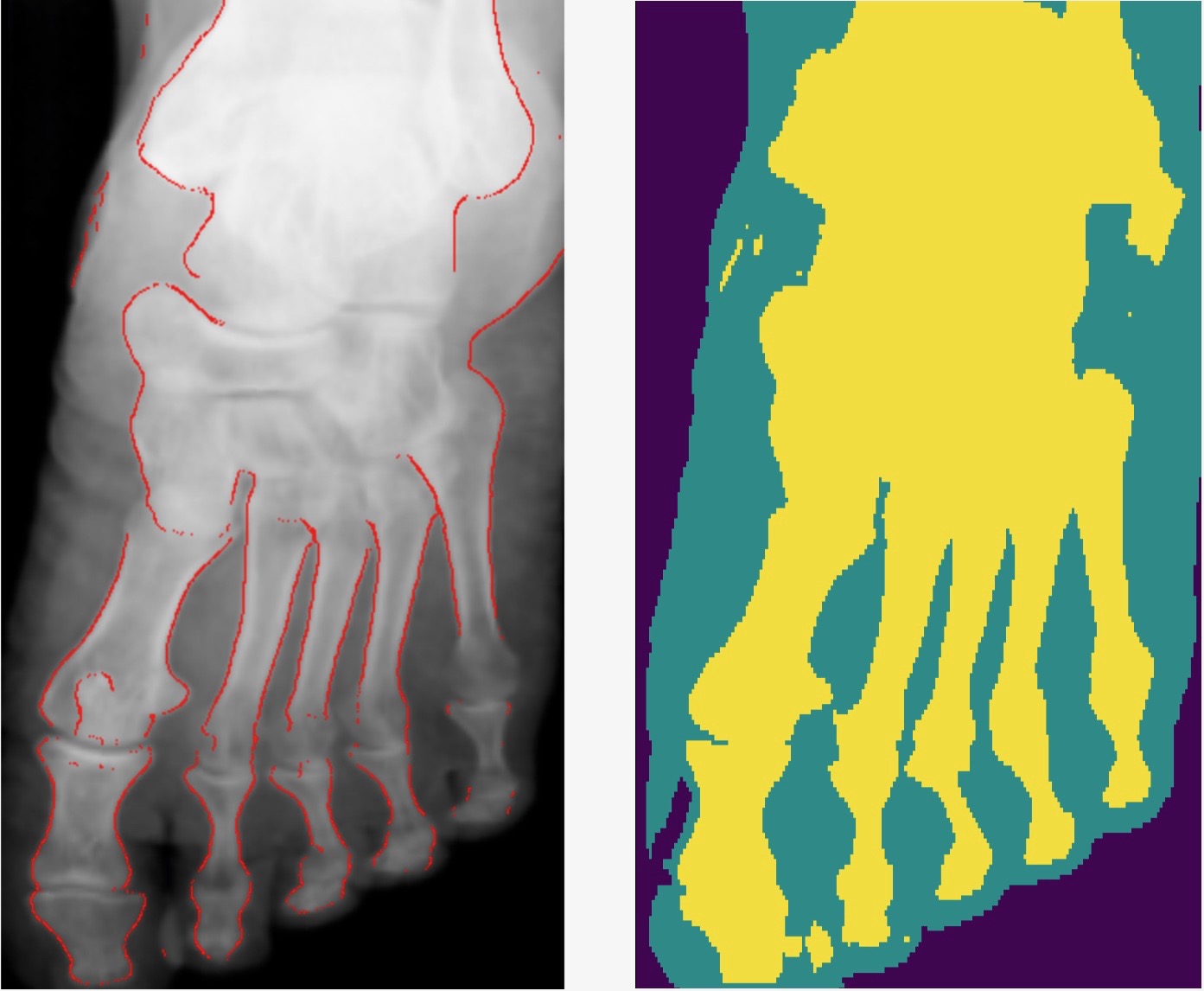}
\end{center}
\caption{Comparison of our algorithm, right, with the work of Kazeminia \textit{et al.} \cite{Line_Fluctuations}, left, showing significant improvement in boundary smoothness}
\label{foot}
\end{figure}

We also see a significant improvement when benchmarking against state-of-the-art classical image processing techniques. Figure \ref{foot} shows a comparison between our architecture and the work of Kazeminia \textit{et al.} \cite{Line_Fluctuations}, who built upon that of Bandyopadhyay \textit{et al.} \cite{Entropy_1, Entropy_2}. The XNet segmentation produces smoothly connected boundaries around the bone regions, in addition to differentiating well between bone and soft tissue regions. It should be noted that in producing this output, our algorithm was trained on a set of high resolution \textsc{TIF} images, as produced by the X-Ray scanner, whereas this analysis was run on significantly lower quality \textsc{JPEG} images, meaning our network could not achieve its full potential on this image. Additionally, our training set does not contain any feet viewed from the angle as seen in Figure \ref{foot}, thus demonstrating the generalisability of our network.

Another classical example is the clustering based method presented by Kubilay Pakin \textit{et al.} \cite{Clustering_Approach}, which gives a 92\% accuracy averaged over 14 images belonging to 5 body part classes. This accuracy score is calculated after each of the two free parameters of the algorithm are fine-tuned for each individual image. Our approach obtains a similar accuracy score in a larger, more diverse dataset, generalising well to 19 body part classes without specific tuning. 



\section{Future Work}\label{Future} 

As an improvement to our implementation, we wish to focus on reducing the false positive rate at the post-processing stage. Additionally, there have been several attempts at refining segmentation mask outputs, such as using conditional random fields \cite{Conditional_Random_Fields}. Such methods could be applied to our work to further improve boundary smoothness, and reduce the likelihood of false islands appearing in the mask. Another approach could be to train an adversarial network, along with XNet, to detect inconsistencies between the network generated segmented maps and the ground truth. However, such an implementation would significantly increase the training time of the network, while also increasing the risk of falling into local minima during training.

Furthermore, it would be interesting to  diversify the classification categories to identify different bone types or tissue materials. Identification of such differences, particularly in the tissue region, is of great importance to the medical field since tissue abnormality detection is a highly non-trivial task when using X-Ray images.



\section{Acknowledgements}

The authors are part of the Centre for Doctoral Training in Data Intensive Science at Durham University supported by STFC grant number ST/P006744/1. The authors are grateful to IBEX Innovations Ltd. \cite{IBEX} for hosting them and providing data, and to Kazeminia \textit{et al.} \cite{Line_Fluctuations} and Bandyopadhyay \textit{et al.} \cite{Entropy_2} for providing images used in their work. This work made use of the facilities of the Hamilton HPC Service of Durham University. This work also used the DiRAC Data Centric system at Durham University, operated by the Institute for Computational Cosmology on behalf of the STFC DiRAC HPC Facility (\url{www.dirac.ac.uk}). This equipment was funded by BIS National E-infrastructure capital grant ST/K00042X/1, STFC capital grant ST/H008519/1, and STFC DiRAC Operations grant ST/K003267/1 and Durham University. DiRAC is part of the National E-Infrastructure. The authors are grateful to the above organisations for their support.
\bibliography{bib} 
\bibliographystyle{spiebib} 
\end{document}